\documentclass{article}

\usepackage{microtype}
\usepackage{graphicx}
\usepackage{subfigure}
\usepackage{booktabs} % for professional tables

\usepackage{graphics} % for pdf, bitmapped graphics files
\usepackage{amsmath} % assumes amsmath package installed
\usepackage{amssymb}  % assumes amsmath package installed

\usepackage{mathtools}
\usepackage{color}
\def\mathbi#1{\textbf{\em #1}}

\usepackage{hyperref}

\usepackage[accepted]{icml2018}

\icmltitlerunning{Bayesian Optimization for Dynamic Problems}

\begin{document}

\twocolumn[
\icmltitle{Bayesian Optimization for Dynamic Problems}



\icmlsetsymbol{equal}{*}

\begin{icmlauthorlist}
\icmlauthor{Favour M.~Nyikosa}{ox} 
\icmlauthor{Michael A.~Osborne}{ox} 
\icmlauthor{Stephen J.~ Roberts}{ox}
%\icmlauthor{Aeiau Zzzz}{equal,to}
%\icmlauthor{Bauiu C.~Yyyy}{equal,to,goo}
%\icmlauthor{Cieua Vvvvv}{goo}
%\icmlauthor{Iaesut Saoeu}{ed}
%\icmlauthor{Fiuea Rrrr}{to}
%\icmlauthor{Tateu H.~Yasehe}{ed,to,goo}
%\icmlauthor{Aaoeu Iasoh}{goo}
%\icmlauthor{Buiui Eueu}{ed}
%\icmlauthor{Aeuia Zzzz}{ed}
%\icmlauthor{Bieea C.~Yyyy}{to,goo}
%\icmlauthor{Teoau Xxxx}{ed}
%\icmlauthor{Eee Pppp}{ed}
\end{icmlauthorlist}

%\icmlaffiliation{to}{Department of Computation, University of Torontoland, Torontoland, Canada}
%\icmlaffiliation{goo}{Googol ShallowMind, New London, Michigan, USA}
%\icmlaffiliation{ed}{School of Computation, University of Edenborrow, Edenborrow, United Kingdom}
\icmlaffiliation{ox}{Machine Learning Research Group and Oxford-Man Institute of Quantitative Finance, University of Oxford, Oxford, United Kingdom}

\icmlcorrespondingauthor{Favour M.~Nyikosa}{favour@robots.ox.ac.uk}
%\icmlcorrespondingauthor{Eee Pppp}{ep@eden.co.uk}

\icmlkeywords{Machine Learning, ICML}

\vskip 0.3in
]



\printAffiliationsOnly{} 


\begin{abstract}
We propose practical extensions to Bayesian optimization for solving dynamic problems. We model dynamic objective functions using spatiotemporal Gaussian process priors which capture all the instances of the functions over time. Our extensions to Bayesian optimization use the information learnt from this model to guide the tracking of a temporally evolving minimum. By exploiting temporal correlations, the proposed method also determines when to make evaluations, how fast to make those evaluations, and it induces an appropriate budget of steps based on the available information. Lastly, we evaluate our technique on synthetic and real-world problems.
\end{abstract}

\section{Introduction} \label{sec: intro}

Global optimization problems arise in a wide range of applications. However, there is a class of global optimization problems where the objective function is unknown, is expensive to evaluate and we have no access to its derivative information. Bayesian optimization (BO) \cite{bo7} is a popular approach for solving this type of problem. It models the latent objective function using a surrogate model and uses an auxiliary optimization process to sample it sequentially. A popular surrogate model is a Gaussian process (GP) prior because of its ability to handle uncertainty about the latent objective function. Bayesian optimization has been used extensively in applications such as robotics \cite{bo13}, algorithm configuration \cite{bo1}, sensor selection \cite{bo6}, adaptive Markov chain Monte Carlo (MCMC) \cite{boMCMC}, localization \cite{CarpinRKR15} and probabilistic programming \cite{wood1}.

There is another class of problems to consider where the objective function, or its constraints, evolve with time. Such complex problems naturally arise in dynamic environments. This also introduces the need to select when to evaluate a function of time in order optimize it. So we address two things: \textit{what} evaluation of the objective function to make and \textit{when} to do so.

% discuss meta-heuristics
There are other reasons why studying this type of problem is important. As we are currently deluged with information from all aspects of society, the resulting problems are usually very complex and can be difficult to solve or automate, especially for tasks that arise naturally. As a result, it is often much easier to produce \emph{meta-solutions} and \emph{meta-heuristics} that provide sufficiently good methods for these problems. The main benefit of working on dynamic optimization problems (DOPs) is that the resulting solutions are meta-heuristics for many kinds of difficult real-world problems. It is, therefore, necessary to study dynamic optimization problems of this nature.

\textbf{Related Work:} Nature-inspired population-based optimization algorithms were the first kind of solutions for DOPs. Examples of such algorithms include evolutionary algorithms (EAs) \citep{dop1}, particle swarm optimization (PSOs) \citep{dop6}, ant colony optimization (ACOs) \citep{dop7}, and immune-based methods \citep{dop8}, among others. These methods have mostly been tested on synthetic problems. Many test benchmarks have been created, most notably the moving peaks benchmark \citep{dop9}, which is the most widely used. This benchmark has controllable complexity and degree of dynamism. In the last ten years, there have been some applications to real-world DOPs in aerospace design \citep{dop10}, risk in financial optimization problems \citep{dop12}, path planning for ships and pollution control \citep{dop11}.

While there has been much work done in the area of Bayesian optimization with GPs for continuous spaces, there has been less for dynamic problems. In \citet{bo28}, they propose a sequential BO algorithm with bandit feedback that allows for the reward function to vary with time. They do this by modelling the reward function with a GP prior whose evolution follows a simple Markov model. They tackle the issue of identifying and discarding stale data samples of the time-varying reward function as a way of adapting to its changes. The rationale is that older information may be stale and therefore misleading. Their algorithm outperforms classical Bayesian optimization using upper confidence bounds \citep{bo23}. 

\textbf{Contributions:} The problem with the formulation above is that there are many real-world dynamic problems whose evolution follows more complex behaviour than a simple Markov model. In this paper, we use a more general class of models and propose several extensions to Bayesian optimization for solving dynamic problems on continuous spaces that are unknown and expensive to evaluate. 

%\textbf{Note:} While reinforcement learning (RL) \citep{busoniu2008comprehensive} and BO are similar in that they balance exploration and exploitation of an unknown function to achieve their goals, they differ in three fundamental ways. First,  BO does not involve any state. Second, BO seeks to find a global extremum rather than learn a policy to govern an agent’s behaviour. Third, RL is normally used as the primary system for a particular task while BO is often used within larger systems, like for the system's configuration.

\section{Problem Statement} \label{sec: problem}

We define a dynamic optimization problem (DOP) \cite{dop1} as follows:
	\begin{flalign}\label{eq: dop_intro}
	    \text{DOP } = \Big \{ 
	    \begin{split} 
	        \text{ minimize      }  &    f( \mathbi{x} , t)\\
	        \text{ s.t. \: \: \: \:   }  &    \mathbi{x} \in F(t) \subseteq \mathcal{S}, t \in \mathcal{T} \:                                  
	    \end{split}
	     \Big \}
	\end{flalign}
	where:
	\begin{itemize}
	   \item $\mathcal{S} \in \mathbb{R}^{D}$, $\mathcal{S}$ is the search space. 
	   \item $t \in \mathbb{R}$ is the time.
	   \item $ f: \mathcal{S} \times \mathcal{T}  \mapsto \mathbb{R} $ is the objective function that assigns a numerical value ($f(\mathbi{x}, t) \in \mathbb{R}$) to each possible solution ($ \mathbi{x} \in \mathcal{S} $) at time $t$.
	   \item $F(t)$ is the set of all feasible solutions $ \mathbi{x} \in F(t) \subseteq \mathcal{S} $ at time $t$.
	\end{itemize}

A DOP is one where the objective function, or the constraints on how to optimize it, change with time. The most straightforward way of solving this problem is to ignore the dynamics and consider each change as a new optimization problem that we want to solve. This re-optimization strategy may be impractical, as shown in \cite{dop2}. Therefore, the goal of solving a DOP is no longer to find a stationary optimal solution but to have a mechanism for efficiently keeping track of its movement through the solution space. We assume the temporal evolution is an inherent property of the DOP, so an optimization algorithm does not modify it through evaluations. Thus the goal of any solution method is to find the global minimum and keep track of it without having to restart the process. An ideal solution is one that tracks the minimum closely on average within a fixed horizon of iterations. 

\textbf{Bayesian Optimization}: Bayesian optimization (BO) is a sequential design strategy for optimizing an unknown, noisy and expensive objective function. We often assume that we have a way of obtaining expensive samples from this objective function. This operation would usually involve taking a costly or time-consuming physical experiment or simulation.

BO operates in two significant phases. The first phase uses a surrogate model to learn the latent objective function from available samples. A popular surrogate model is a Gaussian process prior \citep{gp1}. The second phase involves finding a suitable point to sample the objective function. This search is performed using some predefined heuristics to get closer to the optimum.  The general strategy is to have heuristics that intelligently and automatically explore and exploit the objective function. In practice, this sampling choice is achieved by performing a secondary optimization of a surrogate-dependent acquisition function $a( \mathbi{x}, t)$. Popular acquisition functions are the improvement-based expected improvement (EI) and optimistic upper confidence bounds (UCB). After this sample point is obtained, the objective function is evaluated at that point, and the process is repeated. Algorithm \ref{alg: bo} delineates the method for an objective function $y = f(\mathbi{x}, t)$. For a detailed overview, see \cite{bo25}.

\begin{algorithm}[t]
    \caption{Bayesian Optimization (BO)}
    \label{alg: bo}
    \begin{algorithmic}
        \FOR {$\boldsymbol{i} = 1,2,... \text{ \{Max iterations\}}$} 
        
        \STATE Train GP model 
        \STATE Calculate $ \{ \mathbi{x}_i, t_i \} = \underset{x,t}{ \text{arg max  } } a( \mathbi{x}, t)$  
        \STATE Query objective function $ \mathbi{y}_i \leftarrow f(\mathbi{x}_i, t_i)$        
        \STATE Augment new point to the data
        \STATE Update current location of optimum
                
        \ENDFOR
    \end{algorithmic}
\end{algorithm}

\textbf{Gaussian Process}: A GP $\mathcal{GP}(\mu, K)$ is a collection of random variables where any finite subset has a joint Gaussian distribution. It describes a prior distribution over functions, and it is completely specified by a mean function $\mu: \mathcal{X} \mapsto \mathbb{R}$, $\mu(\mathbi{x}, t) = \mathbb{E}[ f(\mathbi{x}, t)  ]$ and covariance function (or kernel) $K: \mathcal{X} \times \mathcal{X} \mapsto \mathbb{R}$, $K( (\mathbi{x}, t), ( \mathbi{x}^{\prime}, t^{\prime} ) ) = \mathbb{E}[ ( f(\mathbi{x}, t) - \mu(\mathbi{x}, t )) ( f( \mathbi{x}^{\prime}, t^{\prime} ) - \mu(\mathbi{x}^{\prime}, t^{\prime} ))  ]$, where $K(\cdot, \cdot) \leq 1$ and $\mathcal{X} = \mathcal{S} \times \mathcal{T} $. For a detailed overview, see \cite{gp1}.

%Suppose we have a dataset $\mathcal{D} = [ \{\mathbi{x}_i, t_i \}, y_i ]_{i=1}^{i=N} =  \{ \mathbi{X}, \mathbi{y}\}$ where $y_i = f(\mathbi{x}_i, t_i) + \epsilon$ where $y, t \in \mathbb{R}$, $\mathbi{x} \in \mathbb{R}^{D}$, and $\epsilon \sim \mathcal{N}(0, \sigma_n^2)$ is the noise. Using a GP prior $p(y) = \mathcal{N}( \mathbi{0}, k(\mathbi{x}, \mathbi{x}^{\prime}  ) )$ and a test input $\mathbi{x}^{\star}_t = \{ \mathbi{x}^{\star}, t^{\star} \}$, we can obtain a posterior distribution for the associated target ${y}^{\star}$ with the posterior mean and posterior variance given by:
%	\begin{equation}\notag
%		\begin{aligned}
%	      \mu(\mathbi{x}^{\star}_t) &= K( \mathbi{x}^{\star}_t, \mathbi{X} )  K( \mathbi{X}, \mathbi{X}  )^{-1} \mathbi{y},\\
%	      \sigma(\mathbi{x}^{\star}_t) &= K(  \mathbi{x}^{\star}_t ,  \mathbi{X} ) - K( \mathbi{x}^{\star}_t, \mathbi{x}^{\star}_t )  K( \mathbi{X}, \mathbi{X}  )^{-1} K( \mathbi{X}, \mathbi{x}^{\star}_t ).
%	      \end{aligned}
%	\end{equation}

%Since we are not usually certain about the suitable hyperparameters $\boldsymbol{\theta}$ to use, a common selection criterion is maximising the log marginal likelihood $p( \mathbi{y} \mid \mathbi{X})$ of the GP. For a fully Bayesian treatments, Bayesian Monte Carlo \citep{ghahramani2003bayesian} or slice sampling \citep{bo2} can be performed. 

\section{Model for Dynamic Optimization Problems} \label{sec: dop}

We model a DOP $f(\mathbi{x}, t)$ as a spatiotemporal GP where we assume that the objective function at time $t$ is a slice of $f$ constrained at $t$. As a consequence, this GP model captures correlations in space and time and can facilitate adaptation to temporal changes in the objective function over time. Since the goal is to adapt to the objective function through time, the order that we observe the fixed-time objective functions is determined by how they are ordered along the time axis of $f$.

Thus we take 
		$
			f(\mathbi{x}, t) \sim \mathcal{GP}(\mathbi{0}, \: K (\{ \mathbi{x}, t \},\{ \mathbi{x}^{\prime}, t^{\prime} \} )  \:),
		$ 
		where $ (\mathbi{x}, t) \in \mathbb{R}^{D+1}, $ and $K$ is the covariance function of the zero-mean spatiotemporal GP. We will take $K$ to be stationary, separable and of the form:
		$$ 
			K(f(\mathbi{x}, t), f(\mathbi{x}^{\prime}, t^{\prime})) = K_{S}({\mathbi{x}, \mathbi{x}^{\prime} })  \cdot K_{T}({t}, t^{\prime}), 
		$$ 
		where $K_{S}(\cdot,\cdot)$ and $K_{T}(\cdot,\cdot)$ are the spatial and temporal covariance functions, respectively. Unlike in \citet{bo28} where $K_{T}$ is restricted to be an Ornstein-Uhlenbeck process of the form $\epsilon^{(t_i-t_j)}$, where $\epsilon \in (0,1]$ is called the \emph{forgetting factor}, we consider any suitable stationary covariance function as the temporal kernel. This is similar to the product kernels used in contextual bandits \cite{krause2011contextual}, except that our context is time. Spatiotemporal GPs for time-varying models also arise naturally in kernel regressive least squares (KRLS) methods \cite{van2012estimation}, although their formulations only consider a smaller class of temporal kernels.  

For stationary covariance functions, the common hyperparameters include the noise variance, input and output length-scales. For our model, we will assume the time dimension is noise-free and the noise variance in the covariance function will only correspond to the spatial component.

After training, the learnt input length-scales tell us about the variability of the $f(\mathbi{x}, t)$ in space and time. Given that we are using stationary covariance functions, the time-related hyperparameters such as the characteristic length-scale in the time dimension become critical. If we have enough data for effective training, this hyperparameter can be informative about the rate that the fixed-time functions change over time. For example, if this length-scale is extremely large with respect to the time horizon, it means the objective function is not changing. If it is short, it means the function is changing quickly.

\section{Adaptive Bayesian Optimization} \label{sec: abo}

As previously stated, Bayesian optimization works by using existing data samples of the latent function and GP prior to build a posterior distribution of the function over its domain. This posterior is used to construct an acquisition function that leads the search to the minimum by exploiting and exploring the objective function. This sampling is done iteratively until either the exit conditions are met or the execution budget is exhausted. The description in Section \ref{sec: problem} is usually based on an objective of the form $f(\mathbi{x})$, where the goal is to search the whole domain for the minimizer $\mathbi{x}^{\star}$. However, we are interested in finding the minimum $\mathbi{x}^{\star}$ at associated time $t^{\star}$, so that we can track it effectively. Since we are restricted by the real-world nature of time in our model $f(\mathbi{x}, t)$, we can only sample it at times meeting those requirements. To reflect these real-world restrictions on the domain of $f(\mathbi{x}, t)$, we have to add some constraints to the acquisition function optimization process. 

Because we cannot collect samples in the past, the \emph{lower bound} on the time variable $t$ will be dictated by a notion of the \emph{current} or \emph{present time}. As with standard Bayesian optimization, we will assume there is an initial set of time-tagged samples of the latent function $f(\mathbi{x}, t)$. Let's define the \emph{present time} as the one related to the most recently collected data sample. 

\begin{figure}[ht]
		\begin{center}
			\includegraphics[width=\columnwidth]{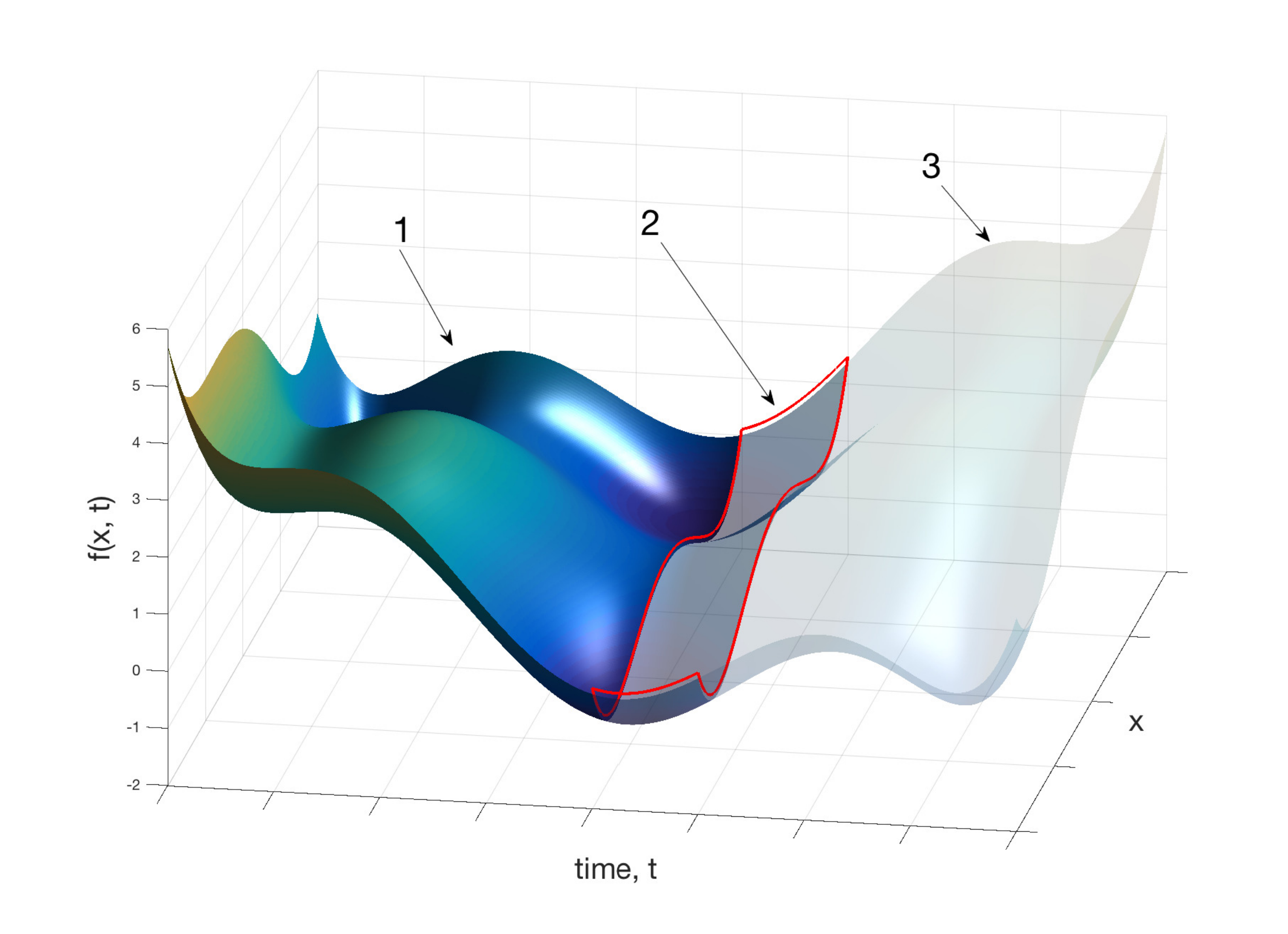}  
		\end{center}	
		\caption{A 2D schematic illustration of $f(\mathbi{x}, t)$ for dynamic problems: Region 1 covers previously observed objective function instances; region 2 shows the bounded region we search at time $t$; and region 3 is the unseen future.}
		\label{fig: space-time}
	\end{figure}

As for the \emph{upper bound}, which dictates how far into the future we are willing to consider solutions for, we ideally want to consider all future times within the fixed time horizon under consideration. This would allow us to solve the optimization problem in one-shot by figuring out where, at some point in the future, we would get the best $\mathbi{x}^{\star}$. However, since we are using a GP prior for $f$, our ability to predict too far from our data samples is quite limited. The capacity to informatively extrapolate with reasonable uncertainty using the posterior distribution is going to be limited to a short distance from the last sample. Specifically, a length-scale away \citep[\S 2.2]{gp1}. In our case, we can only predict well a length-scale way in the time dimension. Therefore, a reasonable upper bound on time would be anywhere within a length-scale distance away from the lower bound. The critical time-related length-scale hyperparameter is learned from the training stage of the GP. This is illustrated in Figure \ref{fig: space-time}.

Algorithm \ref{alg: abo1}, which we call adaptive Bayesian optimization (ABO), shows the resulting adaptive method which changes the bounds and box constraints for optimizing the acquisition function at every iteration based on the learnt time-related length-scale hyperparameter. 
%The sampling strategy from algorithm \ref{alg: abo1} produces samples that are located on one side of the $\mathbb{R}^{D+1}$ space bordered by a line (or hyperplane) corresponding to the time of the previously collected sample. Figure \ref{fig: oneside} illustrates this situation for $\mathbb{R}^{2}$.
	\renewcommand{\algorithmicrequire}{\textbf{Input:}}
	\renewcommand{\algorithmicensure}{\textbf{Output:}}
		\begin{algorithm}[h]
		\caption{Adaptive Bayesian Optimization (ABO)}
			\label{alg: abo1}
		\begin{algorithmic}[1]
		
			\REQUIRE {Initial data $\mathcal{D}_{0}$, GP prior $f(\mathbi{x}, t) \sim \mathcal{N}(\mathbi{y} \mid \mu, K)$} 	
			\REQUIRE Budget of BO steps $N$, iteration label $i$
			
			\REQUIRE Lower bound time delta $ \delta_t$ 
			
			\REQUIRE Upper bound threshold $\rho \in (0,1]$ 
				
			\ENSURE Minimum's trajectory $\{ \mathbi{x}_{i}, t_{i}, y_{i} \}_{i=1}^{N} $
			
			\FOR {$\boldsymbol{i} = 1,2, ... , N$}
				
				\STATE Train GP model
				
				\STATE Set current time $t_c$ from last collected sample
				
				\STATE Set time length-scale $l_{t}$
				
				\STATE Set bounds ${F}(t_i)$ for feasible set $${F}(t_i) = \{ \: \mathcal{S} : (t_{c} +\delta_t) \leq t \leq ( t_{c} + \rho \: l_t ) \: \}$$
				
				\STATE $\{ \mathbi{x}_{i}, t_{i} \} = 
				\underset{ \{ \mathbi{x}, t\} \: \in {F}(t_i) }{\text{arg max } } 
				a( \mathbi{x}, t \mid \mathcal{D}_{i} )$
				
				\STATE  $ y_{i} \leftarrow f(\mathbi{x}_i,{t}_i)$, query objective function
				 				
				\STATE  $\mathcal{D}_{i+1} = \mathcal{D}_{i} \cup \{ \mathbi{x}_{i}, t_{i}, y_{i} \}$, update data
				
		        \STATE Update current location of minimum 
				
			\ENDFOR
		\end{algorithmic}
	\end{algorithm}

The algorithm first trains the Gaussian process model, identifies the current time from the most recently collected data samples and takes note of the learnt time length-scale hyperparameter. The lower bound of the time variable is set to be a $\delta_t$ distance away from the present time $t_c$. We call this distance $\delta_t$ the \emph{lower bound time delta}. This is the minimum length of time into the future we would like to consider solutions for. The upper bound is set to be within a fraction $\rho$ of length-scale from the lower bound. We call this fraction the \emph{upper bound threshold}.   

The method described in Algorithm \ref{alg: abo1} solves for both $\mathbi{x}^{\star}$ and $t^{\star}$ simultaneously, and doing so repeatedly induces the tracking behaviour. However, there are many real-world applications where data samples are collected at fixed intervals. In that case, the constraints at Line 2 become an equality constraints fixed at that the particular time of interest, i.e, the time we know we are taking a sample for the objective function, i.e.
	$
		{F}(t_i) = \{ \: \mathcal{S} : t = (t_{c} +\delta_t) \: \}.
	$

\subsection{Condition for ABO to Work}

For ABO to solve a DOP as described, the function's rate of change over time must be slow enough for us to gather enough samples to learn the relationships in space and time. If the rate of change of the function is faster than we can sample the function, the algorithm learns nothing. The optimal position is where the rate of change is slow enough to capture an evolution pattern.

\subsection{Phases of the Algorithm}

% Talk about the three phases of the algorithm: learning phase, explore ands exploit phase
In its standard form, Bayesian optimization operates by continuously exploiting and exploring the objective function in its search for the minimum. For a dynamic case, this introduces several problems into the procedure. Since the goal is to track the minimum, some explore steps could potentially heavily penalize the algorithm's performance despite the importance of exploration for the learning process. Therefore, it becomes important to critically think about how the balancing of exploration and {exploitation} could be best modified in the cases of dynamic problems, which we discuss next.

\textbf{Initial Design of Data:} One solution arises from the fact that initial data plays a crucial role in training the Gaussian process model before the subsequent sampling procedure. The quality of the information learnt from training the GP determines how fast BO will reach the minimum. Since the sampling choice automatically explores and exploits the surrogate model, good initial datasets facilitate a data efficient way of finding the optimum.

In the case of adaptive Bayesian optimization, the \emph{initial design} affects how well the time length-scale hyperparameter is learnt and also determines the quality of knowledge about the next sampling time of interest. 

For a known budget of BO steps, we can allocate a few of those to obtaining these initial samples for the purpose of learning the function well without worrying about optimization. Ideally, this should be a number of steps we are willing to ``throwaway" to aid better tracking at future steps. One way to select these initial samples is via BO steps that explore and exploit the allowable space. Another way is to generate them randomly but place them in a space-filling manner using a Latin Hypercube Design (LHD). This placement would ensure that  the initial data is spread out and covers as much of the allowable domain within those few first steps. 

A mixture of these methods can also be used. This initial budget of steps allocated to \emph{learning} can be further split between LHD and some explore-exploit BO steps that are not scored for any performance but are used to place additional points in the space to aid the optimization. Doing this takes advantage of the benefits of the initial design and the BO heuristics for learning purposes.   

\textbf{Delimiting the Learning Phase:} It would be very helpful to have a way of determining when the GP model of the underlying latent function that is learnt is good enough to facilitate good predictions within the immediate time horizon under consideration at a particular time step $t$. One way to achieve this is by keeping track of the learnt length-scales of the time dimension $l_{t}$ and the changes at every iteration. Given the sequential nature of our algorithm, we expect that the resulting length-scales after training at each stage would not change much if the model has been learnt well enough. Therefore, we can use the rate of change of the learnt length-scales to determine if this has occurred. For length-scales within a time window $w$ to the present time, the rate of change $\Delta_l$ is given by
	$$
		\Delta_l = \frac{ l_{t} - l_{t - w} }{ w }.
	$$
	To determine whether the learning at time $t$ is good enough, we check if 
	$ 
		-r \leq \Delta_l \leq r,
	$ 
	where $r$ the absolute rate of change or the gradient. A value of $ r = 0.1 $ results in reasonably good detection behaviour.
	
\textbf{Flexible Heuristics:} Given that we have a mechanism for knowing when the model of the objective function has been learnt well enough for predictions with acceptable quality, this affords us the opportunity to vary the mode of heuristics we use during the optimization process. The ideal case for the heuristics is for then to explore and exploit the objective function. However, for the purposes of tracking the minimum, this might not always be ideal. 

A simple alternate strategy is to purely exploit the learnt objective function when we detect that we have learnt it decently enough and switch back to an explore/exploit mechanism if that condition changes. We can use the expected loss or lower confidence bounds acquisition function for exploring/exploiting and the minimum mean function for pure exploitation. That way, we decrease the likelihood that an exploration step selecting a sample far away from a minimum we want to track in steps where we can afford to purely exploit. 

However, it must be stated that the automatic explore/exploit is the best strategy to use in the long run and in cases where there is not much information about the objective function. These prescribed modifications for flexible heuristics are only useful if the assumptions about the model being learnt ``well enough'' are true, and they only force greedy behaviour when we can afford it. 

The result is that an exploit and explore strategy will balance the two criterion for the purpose of finding the minimum while the flexible heuristic strategy will result in greedy behaviour whose success will be contingent on whether the latent function is indeed being learnt well. 

%An illustration of the described phases of the algorithm is shown in Figure \ref{fig: phases}.
%	\begin{figure}[h]
%		\centering
%		\includegraphics[width=\columnwidth]{img/phases.pdf}
%		\caption{Illustration of phases during the adaptive optimization process}
%		\label{fig: phases}
%	\end{figure}

\section{Why ABO Works} \label{sec: abo-why}

% in practice
As stated previously, standard Bayesian optimization operates by sequentially exploring and exploiting the latent objective function $f$ with the aim of finding the extremum, and it has been shown to perform well in experiments \cite{gpgo1, bo2, bo25, calandra2016bayesian}.

% theoretical analysis
There have also been studies on the convergence and theoretical analysis of BO with different heuristics. For all these studies, the algorithmic behaviour depends on the GP prior $p(f)$ used to model $f$ and how it gains information about the objective function during the sequential sampling process. For studies on improvement-based acquisition functions, see \cite{vazquez2010convergence, grunewalder2010regret, bull2011convergence}, and for optimistic heuristics, see \cite{bo23}.

% why ABO works
Given how standard BO works, the theoretical analysis shows the effect of the prior on the exploration and exploitation of $f$. A prior that appropriately captures the family of functions that the $f$ belongs to ensures that exploration and exploitation are appropriately balanced for fast convergence. 

Standard BO performs the search for the extremum throughout the domain of $f$. For DOPs, however, the temporal evolution places restrictions on which part of the function domain can be accessed and evaluated at any point in time. 

This domain restriction is addressed by how ABO estimates the feasible set $\mathcal{S}$ from the posterior. This feasible set is selected such that the Bayesian updates from previous steps provide the most informative posterior distribution over the set $\mathcal{S}$. This means that we exclude parts of the domain where the posterior distribution reverts to the prior and where the posterior uncertainty would likely be too high to be informative due to being too far into the future. As a result of these restrictions, the sequential updates to the GP model during ABO provide the same effect as the Bayesian update in the standard case. This is because there is guaranteed to be a minimizer within $\mathcal{S}$ (by definition of a DOP), and thus ABO will behave similarly to standard BO for the relevant GP priors and sampling heuristics.

This observation is supported by the theoretical analysis in \cite{bo28} for the time-varying optimistic case. They characterize a relationship between the forgetting factor $\epsilon$, whose role is akin to the time length-scale, and the time horizon of interest. The consequence is that if the function does not change too drastically as we informatively sample it, we will be able to track the minimum within some provable bounds (see \cite{bo28}).  

%\section{A Note for Practitioners}\label{sec: design-pattern}
%
%To implement ABO in an existing BO codebase, simply include box constraints to the acquisition function optimization.  The constraints are dependent on the length-scales learnt from the previous Bayesian update so you would need to have a way of extracting the relevant length-scales. Allow for some initial iterations where the optimization is performed at a fixed frequency before beginning to optimize for time. 

\section{Experiments} \label{sec: experi}

We have compared the results of our method with alternatives using tests on an extensive set of standard problems and one real-world example. The competing methods are standard Bayesian Optimization (BO), time-varying bandits (TVB) \cite{bo28}, Covariance Matrix Adaptation Evolution Strategy (CMA-ES) \cite{hansen2006cma}, Dividing Rectangles (DIRECT) \cite{jones1993lipschitzian} and Particle Swarm Optimization (PSO) \cite{dop6}. The experiments tested how well the algorithms tracked the minimum of a dynamic optimization problem. The BO-based methods (ABO, BO, TVB) used GP surrogates while and the others did not. The surrogate models used by ABO and TVB took time variations into account while CMAES, DIRECT and PSO adapted to changes by managing the diversity of their evaluations to be reactive to changes of the objective function. We use two versions of our algorithm which we will call ABO-f, which operates at fixed time intervals, and ABO-t, which also calculates when to make an evaluation. Because we assume that the DOPs are expensive to evaluate, we limit the number of allowed evaluations during the optimization process. This limit is a natural stopping criterion. All methods used the same budget of function evaluations during the experiments.

\textbf{Implementation Details:} 
%All the methods were implemented in MATLAB. The GPs were implemented using the GPML codes \cite{rasmussen2010gaussian}. We also used the PSO and DIRECT methods from the Optimization Toolbox, and the CMAES implementation that is freely available from the author's website.
Each method used its recommended default settings except the number of initial evaluations for starting configurations. The BO-based techniques used the lower confidence bound (LCB) heuristic. A parallelized hybrid PSO algorithm performed the optimization of the acquisition functions, where BFGS ran after PSO terminated. Additionally, only two initial samples were generated using an LHD within the allowed region for the initial data. Squared exponential (SE) spatiotemporal kernels were used for BO and ABO, except where stated, while TVB used the Mat\'ern $\frac{1}{2}$ temporal kernel and an SE spatial kernel. The input and output length-scales were estimated using maximum marginal likelihood to make the testing procedure cheaper to perform over many time-steps and experimental batches. We repeated the experiments ten (10) times to obtain the average performance on each problem.

\textbf{Metrics for Assessing Performance:} For static optimization problems, reporting the \emph{best solution} achieved by an algorithm is often enough to assess its quality and for comparisons. Sometimes, additional cost metrics such as time and memory are often considered. However, when studying dynamic problems, choosing a performance metric is not straightforward because there is not only an interest in the quality of the solution, but also in how the algorithms adapt to evolving objective functions.
%For example, an algorithm may initially find a good solution but fail to keep track of it as time progresses. Alternatively, another algorithm can be near-optimal throughout the execution. The former algorithm would perform better than the latter if the \emph{best solution} is used as a metric when users would prefer the second algorithm. The issues highlighted above and the complexity of dynamic problems make creating evaluation criteria a research area in itself, and many metrics have been proposed in the literature \cite{dop1}. 

\begin{figure}
			\begin{center}
				\includegraphics[width=\columnwidth]{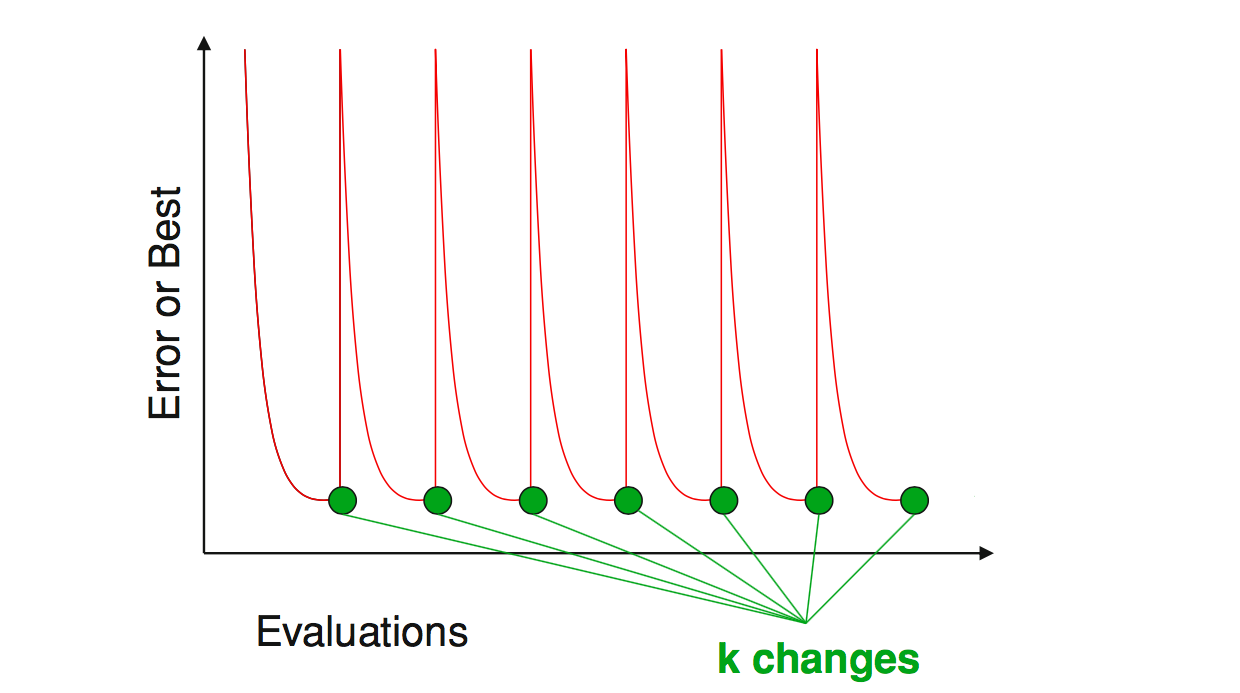}  
			\end{center}	
			\caption{Illustration of how offline-performance works. Source: \cite{dop1} }
			\label{fig: metrics}
		\end{figure}

We are interested in assessing how well an algorithm tracks the minimum on \emph{average}. For this, we will use the offline-performance (B) metric that is defined as
	\begin{equation}\label{eq: gap}
		B(T) = \frac{1}{T}\sum^{T}_{t=1} b_t, 	
	\end{equation}
	%\; \; \;  \text{ and }  \; \; \; E(T) = \frac{1}{T}\sum^{T}_{t=1} e_t,
	where $b_t = \text{min} \{ b_i : i = t, t-1, ...,t-w \}$, the best solution within a window $w$ of iteration $t$ (we use $w=5$). The way offline-performance works is depicted in Figure \ref{fig: metrics}. The trajectory of the best solution is plotted against the iterations. The small circles represent the best solution found before a change occurred. It is assumed that not every evaluation of the objective function contributes to the algorithm's performance. This is particularly useful in the case of BO because some its steps might be exploratory and may not necessarily lead to good solutions, albeit being very beneficial to the algorithm in the long run. Offline-performance does not penalize such steps.

\textbf{Standard Test Objective Functions:}  The first set of experiments were performed on a diverse set of multidimensional test functions (see Table \ref{t: standard}). As a class of dynamic optimization problems, their evolution over time is smooth. For each function, we randomised which dimension was used as the time variable $t$ with the rest being treated as spatial variables for every instance of the batch experiments. 

	\begin{table}[h]
			\caption{Standard test objective functions.}
			\label{t: standard}
			\begin{tabular}{  c  c  c  }
			    \hline
			    {Function} & {Function Name} & { Test Region } \\ 
			    \hline
			    6C & 6-hump Camelback & $[-3, 3] \times [-2,2] $ \\
			    %\hline
			    Br & Scaled Branin & $[0,1]^2$  \\ 
			    %\hline
			    %Col & Colville & $[-10,10]$ & $[-10,10]^{3}$ \\
			    %\hline
			    G-P & Goldstein-Price & $[-2,2]^2$ \\
			    %\hline
			    Gr & Griewank & $[-5,5]^{D}$ \\
			    %\hline
			    H3/6 & Hartmann 3/6 & $[0,1]^{D}$  \\ 
			    %\hline
			    Sh  & Shekel  & $[0,10]^{4}$  \\
			    %\hline
			    Sty2/7 & Syblinsky-Tang 2/7 & $[-5,5]^{D}$ \\ 
			    \hline
	    \end{tabular}	
	\end{table}
	
The results of the experiments are shown in the top part of Table \ref{t: results}. ABO-f and ABO-t performed better (or tied with) the best out of the algorithms tested on most of the test problems. For the cases where it was not the best, like for Hartmann and Shekel functions, it was due to the behaviour of the minimum over time. While the value at the minimum changed over time for those functions, its location did not vary much. As a result, methods that do not consider time variations benefited from exploiting information gained from assuming a static objective function. ABO's exploitation in order to assess how the objective function was varying caused the slightly lower performance in those cases.

\begin{table*}[ht]
	\centering
	\caption{The mean of offline-performances (B) for various optimization methods on various dynamic optimization problems. Numbers highlighted in bold are the best non-ABO-t results for the relevant problem. The last column shows the percentage difference between the iterations used by ABO-t and those used by the other algorithms operating at fixed time intervals.}
	\label{t: results}
	
	%\vskip 0.001in
	
	\begin{tabular}{ c c  c  c  c  c  c | c  c }
		\cline{1-9}
		& {BO} & {CMAES} & {DIRECT} & {PSO}   & {TVB} & {ABO-f} & {ABO-t} & {ABO-t Iters \%-Diff}\\ 
		
		\cline{1-9}
		
		\multicolumn{1}{  c  }{ 6C } &  
		                                 \multicolumn{1}{ c }{0.60} & {0.72} & {0.86} & {1.42} & {0.25} & {\textbf{-0.09}} & {{0.14}} & {-56\%}\\ % {73} & {32}
		
		\multicolumn{1}{  c  }{ Br }                      &
		                              \multicolumn{1}{ c } {-0.69} & {-0.62} & {0.36} & {1.70} & {-0.74} & {\textbf{-0.89}} & {{-0.87}}  & {-16\%}\\ % {41} & {26} 
		
		\multicolumn{1}{  c  }{ G-P } &  
		                                 \multicolumn{1}{ c }{3952} & {6311} & {4686} & {29543} & {16908} & {\textbf{1130}} & {{727}} & {-40\%}\\ %{35} & {21} 

		\multicolumn{1}{  c  }{ Gr } &  
		                                 \multicolumn{1}{ c }{1.11} & {0.61} & {0.88} & {0.82} & {0.57} & {\textbf{0.37}} & {{0.21}} & {+13\%}\\ % 31} & {35}  
		
		\multicolumn{1}{  c  }{ H3 } &  
		                                 \multicolumn{1}{ c }{\textbf{-3.63}} & {-1.18} & {-2.95} & {-3.10} & {-2.50} & {-3.31} & {{-3.77}}  & {+92\%}\\ % {26} & {{50}}
		
		\multicolumn{1}{  c  }{ H6 }                        &
		                              \multicolumn{1}{ c }{\textbf{-2.95}} & {-1.77} & {-1.57} & {-2.05} & {-1.33} & {-2.32} & {{-1.33}} & {-60\%}\\ % {46} & {17}
		
		\multicolumn{1}{  c  }{ She } &  
		                                 \multicolumn{1}{ c }{\textbf{-5.18}} & {\textbf{-5.18}} & {\textbf{-5.18}} & {\textbf{-5.18}} & {{-0.35}} & {{-5.10}} & {{-3.67}}  & {-33\%} \\ % {31} & {24} 

		\multicolumn{1}{  c  }{ Sty2}                        &
		                              \multicolumn{1}{ c  }{-30.2} & {-32.5} & {-44.7} & {-43.4} & {9.9} & {\textbf{-44.9}}  & {{-58.0}} & {-76\%} \\ % {91} & {39}
		                               
		\multicolumn{1}{  c  }{ Sty7 }                        &
		                              \multicolumn{1}{ c  }{-116.1} & {-119.4} & {-66.4} & {-96.9} & {-89.4} & {\textbf{-158.9}} & {{-110.6}} & {-69\%} \\ % {91} & {28} 
		\cline{1-9}                              
		                
		\multicolumn{1}{  c  }{ MPB-1 }                        &
		                              \multicolumn{1}{ c  }{-4.8} & {-23.8} & {-6.7} & {-6.0} & {-25.7} & {\textbf{-29.0}} & {{-16.4}}  & {-27\%} \\   % {230} & {169}                        
		                         
		\multicolumn{1}{  c  }{ MPB-2 }                        &
		                              \multicolumn{1}{ c  }{-0.06} & {-0.03} & {-0.05} & {-0.01} & {-0.01} & {\textbf{-0.64}}
 & {{-0.19}}  & {-16\%} \\     % {400} & {336}                      
		                               
		\multicolumn{1}{  c  }{ Intel }                        &
		                              \multicolumn{1}{ c  }{-1.4e-05} & {-1.5e-73} & {2.2e-3} & {4.3e-10} & {{-1.0e-06}} & {\textbf{-9.7e-3}} & {{-1.0e-2}}  & {+40\%}\\ % {90} & {126}                                           

		\cline{1-9}
		
	\end{tabular}
		
\end{table*}	

The last two columns of Table \ref{t: results} show results related to ABO-t, which also searches for suitable times to make evaluations. Instead of a budget of evaluations, the termination condition for ABO-t was when it reached the end of the time horizon. The columns show the offline-performance of ABO-t and the percentage difference between the iterations used by ABO-t and those used by the other algorithms operating at fixed time intervals. The results show that ABO-t performed well on most of the problems with fewer iterations. In some cases, it used more iterations than the fixed interval case which demonstrated its ability to select evaluation times in a manner that improved tracking performance.

\begin{figure}[ht]
			\begin{center}
				\includegraphics[width=\columnwidth]{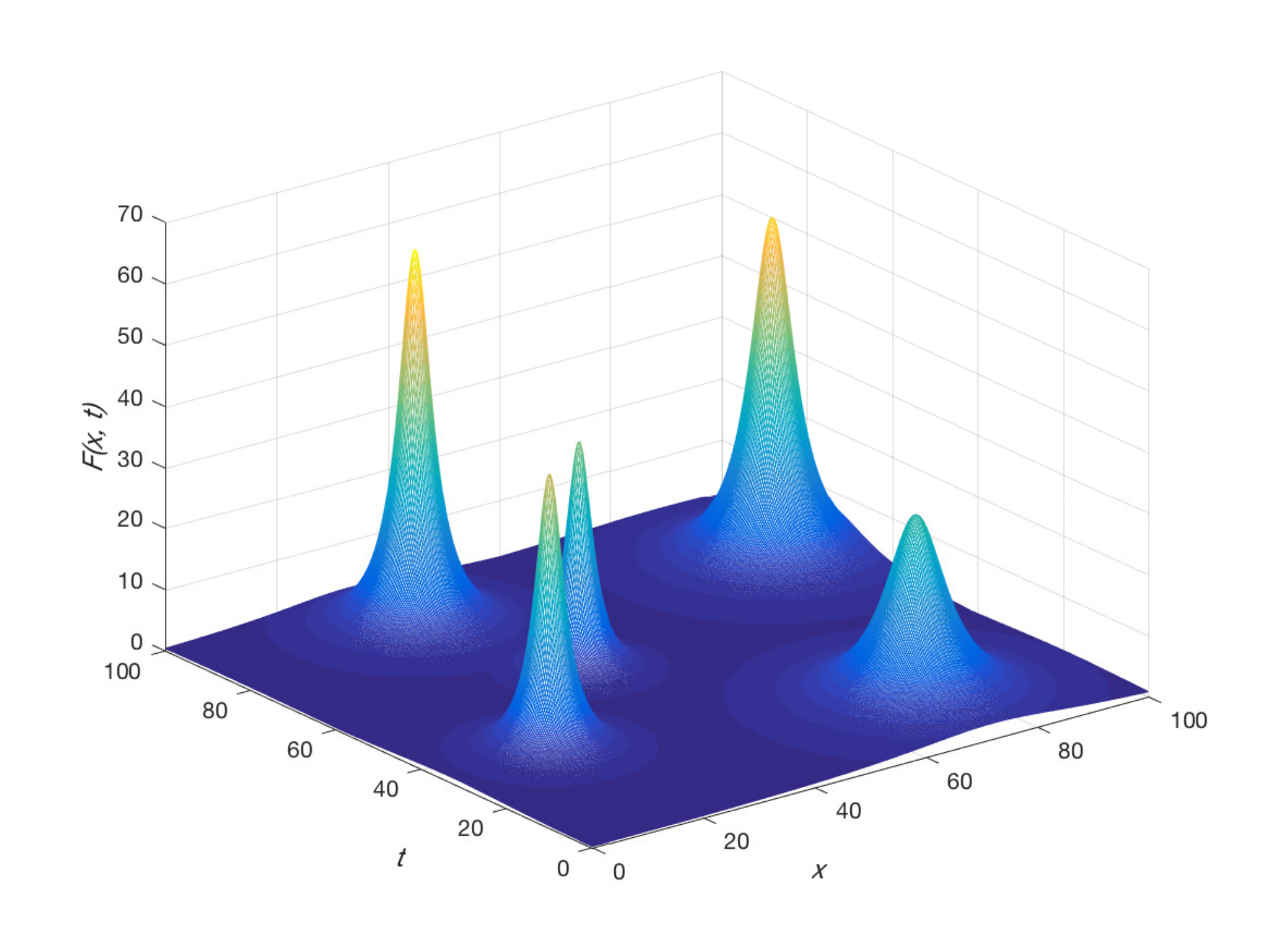}
			\end{center}	
			\caption{Illustration of the moving peaks benchmark with $d=2$ and $m=5$.}
			\label{fig: mpb}
		\end{figure}

\textbf{Moving Peaks Benchmark:} The Moving Peaks Benchmark (MPB) \cite{dop1} has been used extensively to test dynamic optimization approaches. The problem is designed to cover a large range of dynamically changing fitness landscapes. It consists of a multi-dimensional landscape with a definable number of peaks $m$, where the height, width, and position of each peak are altered slightly every time a change occurs. It is defined as
	\begin{equation}\label{eq: mpb}
		F( \mathbi{x}, t ) = \text{max }_{i=1,...,m} \: \frac{H_i(t)}{ 1 + W_i(t) \sum_{i=1}^d (x_i - X_ij(t))^2 },
	\end{equation}
	where $\mathbi{x} \in \mathbb{R}^d$ is the spatial input, $t$ is time, $X \in \mathbb{R}^{m \times d}$ is the set of $m$ peaks locations of $d$ dimensions, and $H, W \in \mathbb{R}^m$ are vectors storing the height and width of every peak. Figure \ref{fig: mpb} shows an illustrative scenario with $ m = 5 $ peaks and $ d = 2 $ dimensions.

The coordinates, the height $H_i$ and the width $W_i$ of each peak are randomly initialized. At certain times, the height and width of every peak are changed by adding a random Gaussian variable multiplied by a specific ``severity" factor ($h_{sev}$, $w_{sev}$, respectively). The location of every peak is moved by a vector $\mathbi{v}$ of fixed length $s$ in a random direction for $\alpha = 0$ or a direction depending on the previous direction for $\alpha > 0$. Thus $\alpha$ is a correlation coefficient allowing to control whether the changes exhibit a trend or not. 

Given $\sigma = \mathcal{N}(0,1)$ and $\mathbi{r}$ is a randomly drawn vector ($r_i \sim \mathcal{N}(0,1)$), we can describe these changes as
	$$
		H_i(t) = H_i(t-1) + h_{sev} \times \sigma,
	$$
	$$
		W_i(t) = W_i(t-1) + w_{sev} \times \sigma,
	$$
	$$
		X_i(t) = X_i(t-1) + \mathbi{v},
	$$
	$$
		v(t) = \frac{ x_{sev} }{ \mathbi{v}(t-1) + \mathbi{r} } \Big[ (1-\alpha)\mathbi{r} + \alpha \mathbi{v}(t-1) \Big].
	$$

%The complexity of the problem can be augmented by increasing the number of dimensions or the number of peaks and by overlaying the whole landscape with high-frequency noise. 
For our experiments, we performed tests on popular scenarios 1 (10 dimensions) and 2 (50 dimensions) with the settings described in \cite{moser2013dynamic}. The results are shown in bottom part of Table \ref{t: results}. ABO-f performed best on the tests. Since the MPB has abrupt changes in the objective function, these results demonstrate how ABO can capture non-smooth time dynamics. The non-BO methods performed worse than ABO despite being able to solve this problem if allowed more iterations. ABO performed better with fewer iterations because it is more data-efficient. The other BO-based methods do even worse. This is because standard BO does not consider time dynamics and TVB's time variation model is not general enough to capture the dynamics in the MPB.

\begin{figure}[ht]
		\begin{center}
			\includegraphics[width=\columnwidth]{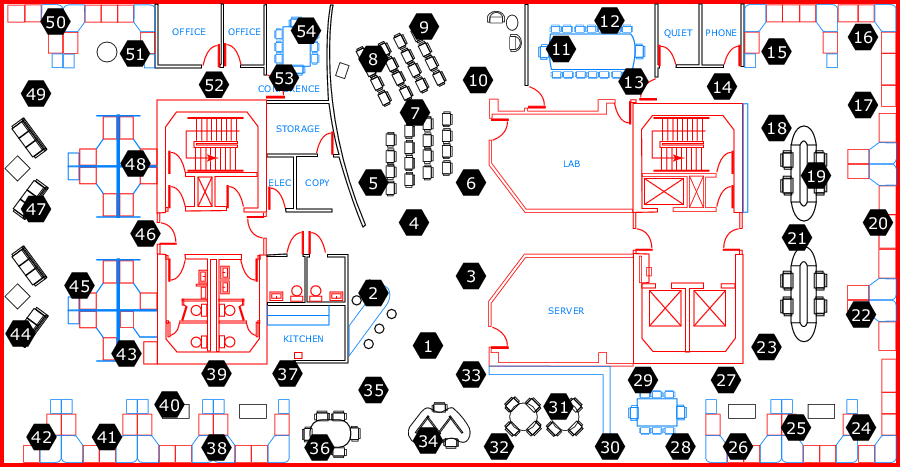}  
		\end{center}	
		\caption{Intel Berkeley laboratory sensor deployment map. Source: \cite{madden2003intel}}
		\label{fig: lab}
\end{figure}

\textbf{Temperature Dataset:} The third set of experiments is on temperature data from 54 sensors at the Intel Research Laboratory in Berkeley, California between February $28$th and April $5$th, 2004, deployed as shown in Figure \ref{fig: lab} \cite{madden2003intel}. The goal was to find which location on the map had the maximum temperature at any given time.

The data contained the sensor location coordinates $x$ and $y$ (in metres relative to the upper right corner of the lab). The experiments used this spatial orientation. The dataset also contained readings collected from these sensors with the schema: \texttt{date, time, epoch, temperature}. Epochs are a monotonically increasing number sequences from each sensor. Two readings with the same epoch number were produced from different sensors at the same time. There were some missing epochs in this dataset.

We considered data over the first $3000$ readings. Due to the large size of the dataset for batch testing, the GP model was trained on $66\%$ of the data and used the learnt hyperparameters for the optimization on the rest of the data sequence. For the non-BO methods, the first $66\%$ of the budgeted steps were not scored but used to configure the algorithms. 

The bottom part of Table \ref{t: results} also shows the outcome of the experiments. ABO had the best mean offline-performance, and the BO-based methods did better than the rest.  For ABO, we used a sum of SE and Mat\'ern $\frac{1}{2}$ covariance functions for both the spatial and temporal components. ABO-t used 40\% more steps than the fixed interval algorithms because the time length-scale hyperparameter was shorter than the predefined fixed time intervals, so ABO-t evaluated the objective function more frequently.

%---------------------------------------------------------------------------------------------------------------------
%	                                                RESULTS AND DISCUSSION
%---------------------------------------------------------------------------------------------------------------------

\subsection{Discussion}

The experiments show that ABO performs well on problems with different kinds of dynamics such as those where the time variations are smooth and those where the changes are abrupt. While the competing methods are good optimization methods under the right conditions, they suffer because they either do not take advantage of time variations or rely on having numerous function evaluations to get good performance. Given that we are looking at problems where function evaluations are expensive, having good performance with fewer evaluations is desirable.

Efficient learning of the DOP model explains the performance of ABO in these experiments. The ability to exploit available prior information and incorporate it into the optimization of the acquisition function demonstrated that the proposed extensions to Bayesian optimization allow for efficient tracking of the minimum. 

Additionally, the exploitation of the temporal relationships to learn how the DOP evolved gave the advantage of being able to determine when to make evaluations, how fast to make those evaluations and induced an appropriate budget of steps based on the available information.

\section{Conclusions} \label{sec: concl}

We have proposed practical extensions of Bayesian optimization for dynamic problems that are latent and expensive. It has clear advantages over previous work as demonstrated by experiments on synthetic and real-world problems. There is an additional benefit of being able to schedule optimal evaluations automatically. Our formulation of the DOP as a spatiotemporal GP proved to be an appropriate prior for efficiently tracking the minimum of dynamic problems.

\end{document}